\documentclass{TST}

\usepackage{graphicx}
\usepackage{footmisc}
\usepackage{subfigure}
\usepackage{url}
\usepackage{multirow}
\usepackage{amsmath,amsthm}
\usepackage{amssymb,amsfonts}
\usepackage{booktabs}
\usepackage{color}
\usepackage{ccaption}
\usepackage{booktabs}
\usepackage{float}
\usepackage{fancyhdr}
\usepackage{caption}
\usepackage{xcolor,stfloats}
\usepackage{comment}
\graphicspath{{figures/}}
\usepackage{cuted}  
\usepackage{epstopdf}
\usepackage[numbers,sort&compress]{natbib}

\newtheoremstyle{mystyle}{0pt}{0pt}{\normalfont}{1em}{\bf}{}{1em}{}
\theoremstyle{mystyle}

\newcommand{\upcite}[1]{\textsuperscript{\cite{#1}}}

\newcommand{\nop}[1]{}
\TSTsetup{
type      = {Research Article},
doi       = {xxxx},
title     = {Supervise-assisted Multi-modality Fusion Diffusion Model for PET Restoration},
 author    = {Yingkai Zhang$^\dag$, Shuang Chen$^\dag$, Ye Tian, Yunyi Gao, Jianyong Jiang, and Ying Fu\,\cor{}\\
},
abstract  = {
  Positron emission tomography (PET) offers powerful functional imaging but involves radiation exposure. Efforts to reduce this exposure by lowering the radiotracer dose or scan time can degrade image quality. While using magnetic resonance (MR) images with clearer anatomical information to restore standard-dose PET (SPET) from low-dose PET (LPET) is a promising approach, it faces challenges with the inconsistencies in the structure and texture of multi-modality fusion, as well as the mismatch in out-of-distribution (OOD) data. 
In this paper, we propose a supervise-assisted multi-modality fusion diffusion model (MFdiff) for addressing these challenges for high-quality PET restoration. Firstly, to fully utilize auxiliary MR images without introducing extraneous details in the restored image, a multi-modality feature fusion module is designed to learn an optimized fusion feature. Secondly, using the fusion feature as an additional condition, high-quality SPET images are iteratively generated based on the diffusion model. 
Furthermore, we introduce a two-stage supervise-assisted learning strategy that harnesses both generalized priors from simulated in-distribution datasets and specific priors tailored to in-vivo OOD data. 
Experiments demonstrate that the proposed MFdiff effectively restores high-quality SPET images from multi-modality inputs and outperforms state-of-the-art methods both qualitatively and quantitatively.
},
keywords  = {Positron emission tomography (PET), Diffusion Models, Image Restoration, Multi-modality},
received  = 20xx-xx-xx,
revised   = 20xx-11-xx,
accepted  = 20xx-00-xx,
}

 \address{Yingkai Zhang, Yunyi Gao, and Ying Fu are with the Beijing Institute of Technology, Beijing, 100081, China (e-mail: zhangyingkai@bit.edu.cn; yiiclass@qq.com; fuying@bit.edu.cn).}

 \address{Shuang Chen is with the Research and Development Center of Agricultural Bank of China, Beijing, 100073, China (e-mail: chenshuangxd@foxmail.com).}

 \address{Ye Tian is with the Peking University, Beijing, 100871, China (e-mail: tye@pku.edu.cn).}
 
 \address{Jianyong Jiang is with the Beijing Normal University, Beijing, 100872, China (e-mail: jianyong@bnu.edu.cn).}

\EditorNote{\hspace{-.7em}$\sf{\dag}$ Equal Contribution.}

\EditorNote{\hspace{-.7em}\cor{} To whom correspondence should be addressed.}

\begin{document}

\maketitle

\section{Introduction}
\label{sec:introduction}

\begin{figure}[t]
\centering
\centerline{\includegraphics[width=\columnwidth,scale=0.4]{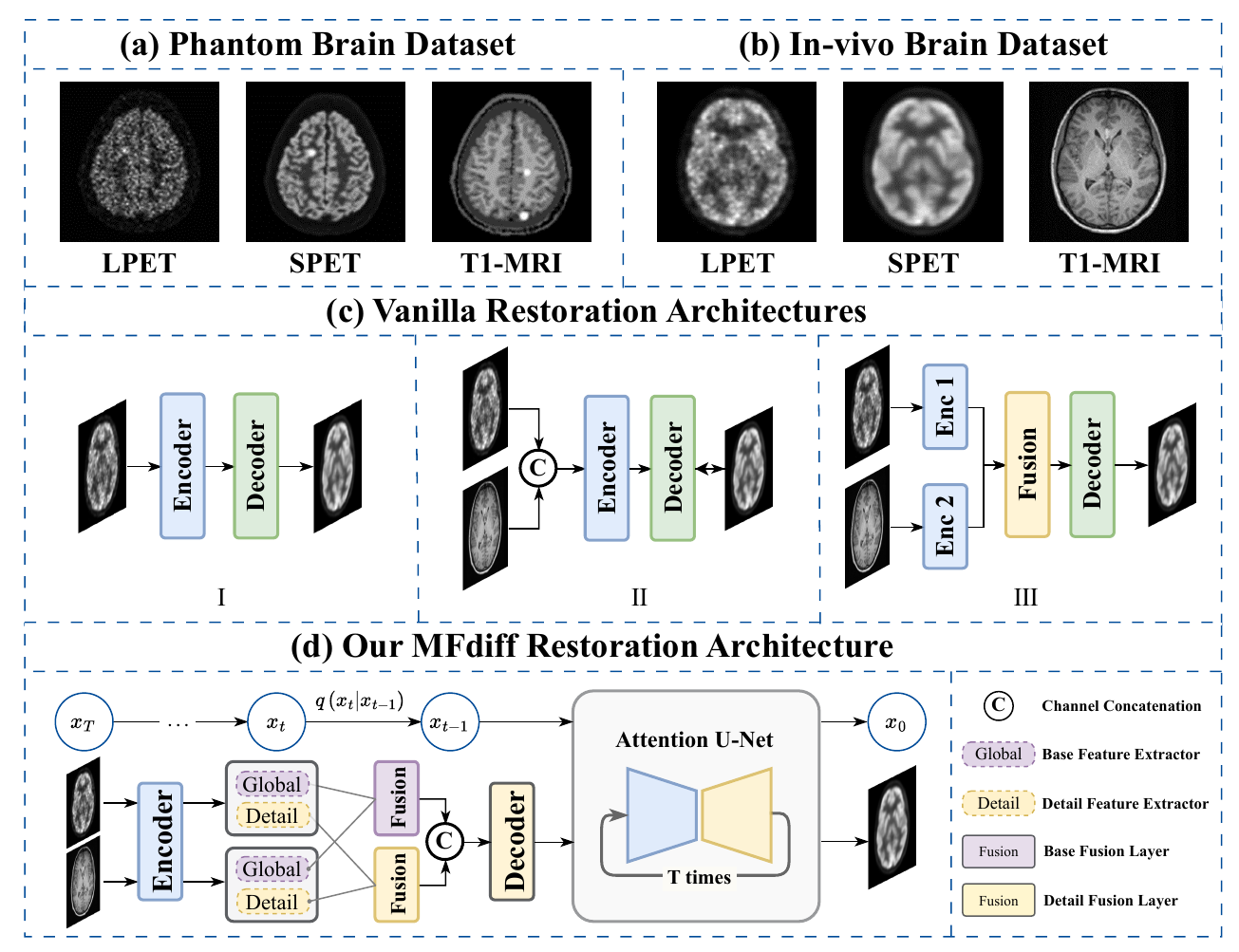}}
\caption{(a-b) Two examples of T1-MRI, LPET, and the corresponding SPET images from the phantom and in-vivo dataset. (c) The categorization of existing PET restoration methods. (d) Our MFdiff restoration method.}
\label{fig:introduction}
\end{figure}

Positron emission tomography (PET) is a powerful molecular imaging technique, crucial in clinical diagnosis and scientific research. It is extensively employed for diagnosing and staging cancer, screening for cardiovascular diseases, researching brain function and neurological diseases, and in drug development~\upcite{alavi2004implications,willmann2008molecular,cherry2018total}. 
Although PET is a highly versatile functional imaging modality, the substantial radiation exposure from PET tracers limits its use and raises significant health concerns. Researchers follow the principle of "as low as reasonably achievable" (ALARA)~\upcite{voss2009alara} when injecting radiotracers, attempting to minimize radiation doses to patients. However, a lower injection dose significantly reduces the number of acquired photon counts, resulting in a lower signal-to-noise ratio (SNR) and compromising PET images, which could potentially impact the final clinical diagnosis~\upcite{liu2021artificial}. Two examples of paired low-dose PET (LPET) and standard-dose PET (SPET) from phantom and in-vivo datasets are shown in Fig.~\ref{fig:introduction}(a-b).

In recent years, benefiting from the continuous development and advancement of algorithms in the field of computer vision~\upcite{tian2023transformer, zhang2024deep,gao2025grayscale,zhang2025unaligned,2024freqfusion,zou2024eventhdr}, various deep learning-based methods have been widely applied in the field of medical image processing~\upcite{huang2021chan,haggstrom2019deeppet,yang2023classification,zhang2023sunet}. They employ end-to-end learning to transform low-dose (LD) into standard-dose (SD) PET images, to restore high-quality PET images. Prior work has primarily focused on LD/SD PET image pairs, but the information available from LD PET images is limited, which presents a bottleneck in SD PET image restoration. The workflow is depicted in \uppercase\expandafter{\romannumeral1} of Fig.~\ref{fig:introduction}(c). With the rising demand for early and accurate disease diagnosis, advanced medical imaging devices such as PET/MR scanners have been widely used in clinical examinations~\upcite{torigian2013pet}. As observed in Fig.~\ref{fig:introduction}(a-b), MR images provide clear anatomical details and superior tissue differentiation, potentially enhancing PET restoration quality. Consequently, several approaches~\upcite{xiang2017deep, da2020micro} have attempted to supply neural networks with anatomical information from simultaneously acquired MR images via additional input channels to aid PET restoration. Compared to using LPET images alone, multi-modality input has proven to significantly enhance the restoration quality of SPET images, as illustrated in \uppercase\expandafter{\romannumeral2}-\uppercase\expandafter{\romannumeral3} of Fig.~\ref{fig:introduction}(c). However, significant data deviations between modalities may result in the inadvertent retention of anatomical and structural information in the restored SPET images, leading to undesirable residual artifacts~\upcite{onishi2021anatomical}. Therefore, we aim to effectively utilize deep semantic features from anatomical guidance images to enhance image quality while carefully avoiding the introduction of extraneous patterns and structural details.

From the perspective of medical image acquisition, large datasets are often unavailable publicly due to concerns related to patient privacy, ethics, and data sharing protocols. In medical imaging, particularly with PET scans, concerns about harmful ionizing radiation make it nearly impossible to acquire significant numbers of paired low-dose and standard-dose data from volunteers, resulting in small-scale PET datasets. Moreover, the complexity of medical imaging processes means that even slight variations in settings can affect the characteristics of acquired PET images, leading to out-of-distribution (OOD) data. 
Factors such as the type and dosage of radiotracers, scan time, hardware, and anatomical changes influence these outcomes~\upcite{accorsi2010improved}.
Data-driven deep learning methods, typically trained in specific scenarios, perform well on in-distribution (ID) data. However, they often face a performance bottleneck due to limited information and cannot directly adapt to out-of-distribution (OOD) data, resulting in comparatively low image quality that challenges the maintenance of clinically applicable performance.
Due to (i) limited dataset size and difficulty in accessing sufficient LD/SD data pairs, (ii) performance degradation of dataset-specific models in handling OOD data, enhancing the generalization capability of models has become a critical challenge in medical imaging applications. 
Consequently, we aim to maintain high restoration quality across specific clinical acquisition variations (e.g., single center, scanner, and tracer) without relying on large-scale paired clinical datasets for each specific scenario.

In this paper, we propose a novel supervise-assisted multi-modality fusion diffusion model (MFdiff) with a multi-modality feature fusion module to restore SPET images from LPET and corresponding MR images acquired through joint PET/MR imaging. The framework is illustrated in Fig.~\ref{fig:introduction}(d) and Fig.~\ref{fig:framework}. 
The multi-modality feature fusion module primarily consists of a dual-branch intra-modality learning (IML) module and a cross-modality aggregation (CMA) module. Initially, the IML module extracts global and detailed features based on the global attention and local convolution from each modality. Subsequently, the CMA module learns an optimized fusion feature representation, which enhances restoration quality and minimizes the impact of modality mismatch. 
Furthermore, we adopt a two-stage supervise-assisted learning strategy that circumvents the need for large-scale paired data in the target clinical domain. Our model is trained on a substantial corpus of paired synthetic data and fine-tuned on a limited amount of domain-specific in vivo data. This strategy allows the network to learn generalized priors from simulated in-distribution datasets while tailoring specific priors to in vivo OOD data. Consequently, the model demonstrates improved adaptability to acquisition-based OOD shifts, thereby enhancing transfer learning efficiency.
The contributions of our work are as follows:
\begin{enumerate}

\item We present a conditional diffusion restoration model based on fusion of the PET/MR to iteratively generate high-quality SPET images.

\item We propose a multi-modality feature fusion module to fully extract and utilize both the specific and shared information of LPET and MR images. We adopt a two-stage supervise-assisted learning strategy with generalized priors and specific priors for high-quality PET restoration.

\item Extensive experiments demonstrate that our proposed method outperforms existing state-of-the-art approaches qualitatively and quantitatively.

\end{enumerate}

The remainder of this paper is organized as follows. Section \uppercase\expandafter{\romannumeral2} reviews related work. The proposed method is detailed in Section \uppercase\expandafter{\romannumeral3}. We introduce the simulated and in-vivo data along with experimental settings in Section \uppercase\expandafter{\romannumeral4}, experimental results in Section \uppercase\expandafter{\romannumeral5}, and conclusion in Section \uppercase\expandafter{\romannumeral6}.

\section{Related Work}
In PET scanning, it is customary to decrease the injection dose of radioactive tracer or reduce the scanning time in order to reduce radiation exposure or increase throughput. These adjustments inevitably lead to increased noise in PET images. Various methods have been proposed to restore SPET images from LPET images, which can be broadly classified into two categories: (i) within-restoration denoising techniques and (ii) post-restoration denoising techniques.

\subsection{Within-Restoration Denoising Techniques}
Within-restoration denoising techniques aim to restore PET images from raw data in list-mode or sinogram domain while simultaneously suppressing noise. To achieve this goal, model-based image restoration methods typically rely on the forward model of the PET system and the designed priors to construct the objective function. For instance, 
Chun \textit{et al.}~\upcite{chun2014alternating} proposed using block-based non-local regularization and employing alternating direction method of multipliers (ADMM) as the solver. 
HÃ¤ggstrÃ¶m \textit{et al.}~\upcite{haggstrom2019deeppet} proposed a direct deep learning based restoration method, named DeepPET, which uses an end-to-end network to learn the mapping of sinogram data to PET images without any imaging system information. It forces the network to learn complex mappings between different domains, which is beyond the range of its abilities, thus harming the performance. To tackle these issues, 
Mehranian \textit{et al.}~\upcite{mehranian2020model} proposed a forward-backward splitting algorithm that unfolds networks into maximum a posteriori (MAP) PET image restoration and utilizes MR images as additional input to guide PET denoising. 
These within-restoration denoising techniques require access to raw list-mode or sinogram data, which is vendor-specific and frequently excluded from large open-access datasets, thereby limiting their practical applications.

\subsection{Post-Restoration Denoising Techniques}
Compared to within-restoration denoising techniques, post-restoration methods that focus on the image domain can benefit from the support of most medical imaging platforms and can be easily and flexibly integrated into existing restoration pipelines. 
Besides, rapid advancements in image restoration tasks~\upcite{li2024latent,li2025noise} within the natural image domain, such as denoising~\upcite{zhang2025real,zou2025calibration,zhang2022guided,wei2021physics}, have significantly accelerated the development of medical image processing (PET).
Initially, filtering-based methods~\upcite{yan2015mri} were used to enhance PET image quality. 
However, these methods are mainly focused on image denoising rather than predicting the SPET image from the LPET image. 
Xiang \textit{et al.}~\upcite{xiang2017deep} were the first to apply CNN to the field of PET restoration. They proposed an auto-context strategy to recover SPET images directly from both LPET images and the corresponding MR images. 
However, methods based on encoder-decoder CNN are typically trained for specific tasks, which limits the generalization of the model across various settings. Furthermore, images processed by CNN frequently exhibit issues such as blurred edges and over-smoothing. 

In recent years, generative adversarial networks (GAN)~\upcite{du2023transformer} have been widely used in the field of image restoration. Compared with methods based on encoder-decoder CNN, they generate images by learning the data distribution, making them more adaptable when dealing with various imaging scenarios. 
Ouyang \textit{et al.}~\upcite{ouyang2019ultra} presented a GAN-based restoration approach for ultra-low-dose amyloid PET images. 
Geng \textit{et al.}~\upcite{geng2021content} proposed a content-noise complementary learning strategy, utilizing two predictors to learn content and noise complementarily, and implemented based on GAN. 
While GAN-based methods can rapidly generate high-quality samples, they often suffer from poor mode coverage, lack of sample diversity, and training instability, leading to issues like mode collapse, vanishing gradients, and convergence difficulties. 

Fortunately, the diffusion model has been proposed to address the limitations of GAN-based methods, offering high sampling quality, extensive pattern coverage, and sample diversity. It excels in learning complex data distributions, 
such as in medical imaging~\upcite{gong2024pet,yoon2024volumetric,shi2024diffusion}, \textit{etc.} 
Previous methods often directly utilize raw multimodal data for image processing. For instance, Gong \textit{et al.}~\upcite{gong2024pet} employed Denoising Diffusion Probabilistic Models (DDPM), using PET and MR images as conditional inputs for denoising. Yoon \textit{et al.}~\upcite{yoon2024volumetric} utilized a Score-based Residual Diffusion Model to learn the residual between paired images. A significant drawback of these direct approaches is that the forward and reverse diffusion processes can inadvertently introduce irrelevant artifacts or lose critical details, potentially compromising clinical diagnostic accuracy. Moreover, the generalizability of diffusion models to OOD PET data from diverse acquisition environments and the effective utilization of such OOD data are critical challenges that must be addressed to develop robust and clinically viable PET restoration methods.

\section{Supervise-assisted Multi-modality Fusion Diffusion Model}

\begin{figure*}[t]
\centering
\includegraphics[scale=0.4]{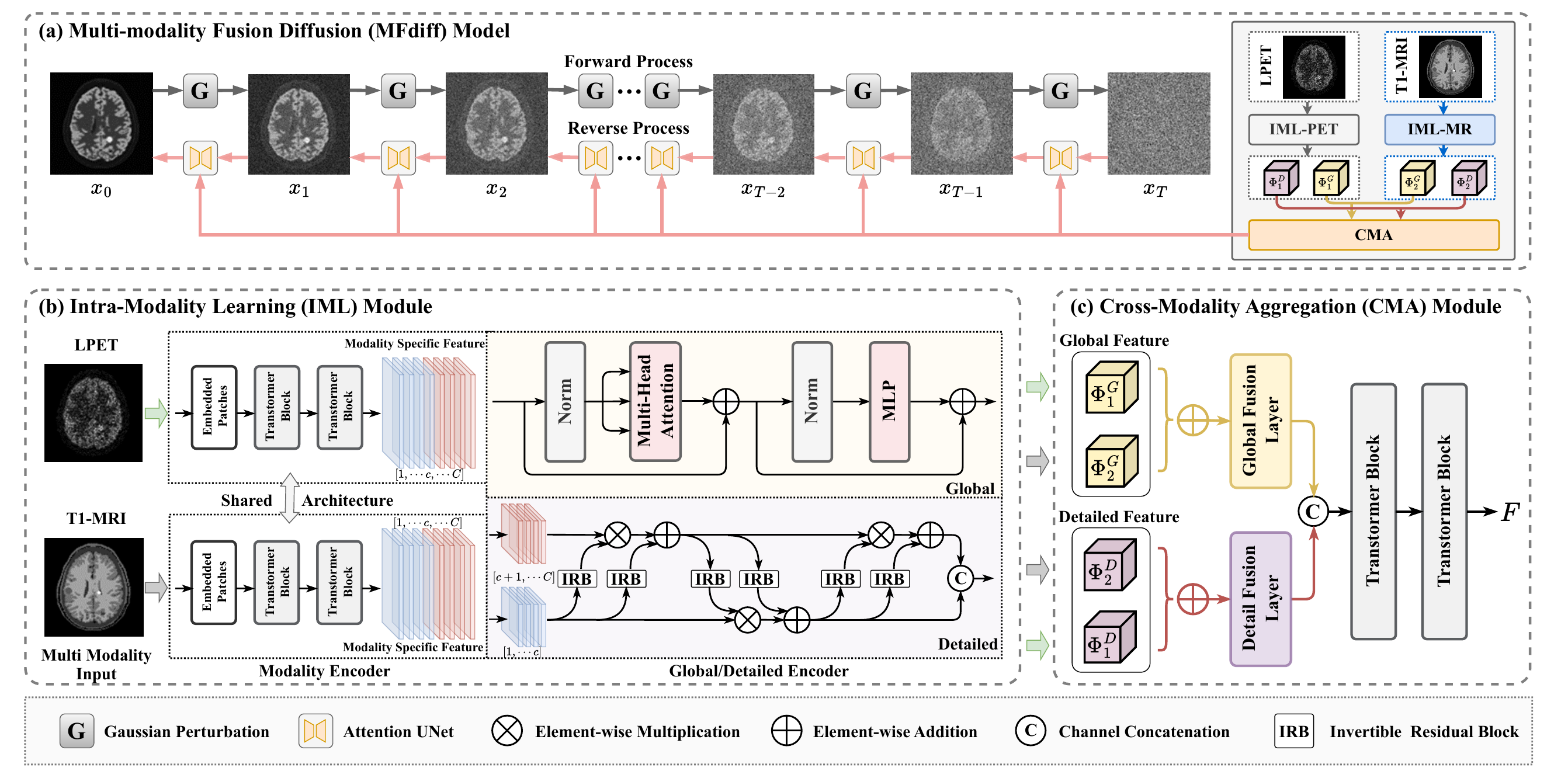}
\caption{Fig. 2 Overview of our supervise-assisted multi-modality fusion diffusion model (MFdiff). (a) provides the framework of MFdiff, which includes two main components: a conditional diffusion restoration module and a multi-modality feature fusion module. The details of the latter module are illustrated in (b) intra-modality learning module with modality ecoder and global/detailed encoder, and (c) cross-modality aggregation module.}
\label{fig:framework}
\end{figure*}

The proposed MFdiff is depicted in Fig.~\ref{fig:framework}(a), including a multi-modality feature fusion module and a conditional diffusion restoration module. 
Suppose that given the LPET $x_{P}\in\mathbb{R}^{H \times W}$ and the reference MRI $x_{M}\in\mathbb{R}^{H \times W}$. $H, W$ are the number of rows and columns. We aim to restore SPET $x_{S}$ by:
\begin{equation}
	x_{S} = f_{\theta}(x_{P}, x_{M}),
\end{equation}
where $x_{S}\in\mathbb{R}^{H \times W}$ and $f_{\theta}(\cdot)$ denotes the multi-modality fusion diffusion model (MFdiff).
Specifically, the multi-modality feature fusion module (detailed in Sec. \ref{sec: Multi-modality Feature Fusion Module}) aims to extract and fuse cross-modality image features by the local and global feature perception capabilities of CNN and Transformer. 
The conditional diffusion restoration module (detailed in Sec. \ref{sec: Conditional Diffusion Restoration Module}) aims to restore high-quality SPET images based on diffusion model.
In Sec. \ref{sec: Two-stage Supervise-assisted Learning Strategy}, we introduce a two-stage supervise-assisted learning strategy to address the lack of real paired data, and OOD data frequently encountered in real clinical settings.

\subsection{Multi-modality Feature Fusion Module}
\label{sec: Multi-modality Feature Fusion Module}

Data disparities exist between PET and MR modalities, accompanied by challenges in effective message interaction. Existing methods, which often directly concatenate data, fail to capitalize on the information from both modalities and may even lead to the underutilization or damage of LPET image. To address these challenges in inter-modal data interaction, we propose the multi-modality feature fusion module shown in Fig.~\ref{fig:framework}(a), which comprises an Intra-Modality Learning (IML) module and a Cross-Modality Aggregation (CMA) module.

\subsubsection{Intra-Modality Learning (IML) Module}
In PET image denoising, anatomical information from the corresponding MR image is expected to enhance the quality of denoised SPET images. However, it is crucial to ensure that detailed information from MR image, which mismatches with PET, is not introduced. Therefore, it is necessary to exploit the potential correlation between the two modalities while capturing modality-specific information, to enhance the features of the target modality and suppress mismatched features from the auxiliary modalities. To achieve this goal, we propose the IML module, as shown in Fig.~\ref{fig:framework}(b). 
Firstly, to extract modality specific features from LPET $x_{P}$ and MRI $x_{M}$, we adopt the `Modality Encoder'~\upcite{zamir2022restormer}. It consists of two independent Transformer encoders with the same shared architecture but different parameters, and can be expressed as:
\begin{equation}\label{eq9}
\begin{aligned}
F_P=f_{\theta_{p}}\left(x_{P}\right), \\
F_M=f_{\theta_{m}}\left(x_{M}\right),
\end{aligned}
\end{equation}
where $f_{\theta_{p}}$ and $f_{\theta_{m}}$ are two modality-specific networks.

Subsequently, to enhance the global features of PET and MR images while distinguishing their respective detailed features, we divide PET and MR images into PET global feature $G_P\in \mathbb{R}^{C \times H \times W}$, PET detailed feature $D_P\in \mathbb{R}^{C \times H \times W}$, MR global feature $G_M\in \mathbb{R}^{C \times H \times W}$, and MR detailed feature $D_M\in \mathbb{R}^{C \times H \times W}$. By reinforcing the consistency of high-level global features between PET and MR, and differentiating the detailed information of each modality, we ensure that the final fused features effectively assist the generation of SPET images. To achieve this, we propose a dual-branch feature extractor (`Global/Detailed Encoder') to further extract global and detailed features from the shallow features of LPET and MRI. The global features $G_P$ and $G_M$ are extracted as:
\begin{equation}
\begin{aligned}\label{eq10}
G_i^{\prime}=MHA(LN(F_i))+F_i, \\
G_{i}=MLP(LN(G_{i}^{\prime}))+G_{i}^{\prime},
\end{aligned}
\end{equation}
where $i=\{P, M\}$ represents the input modality. $MHA$ denotes the multi-head attention, $MLP$ is the multilayer perceptron~\upcite{zamir2022restormer}, and $LN$ is the layer normalization.
The global branch based on Transformer with spatial self-attention extracts global features $G_P$ and $G_M$ from the shallow features $F_P\in \mathbb{R}^{C \times H \times W}$ and $F_M\in \mathbb{R}^{C \times H \times W}$ of each modality. Specifically, the global feature $G_P$ from the low-quality LPET image exhibits a relatively blurry structure and edges, while the global feature $G_M$ from the high-quality MR image captures clear and precise global anatomical information.

As PET image applications are sensitive to fine details, we strive for lossless detail feature extraction to preserve as much valuable information from the original LPET and MR images as possible.
Due to the capability of the invertible neural network~\upcite{dinh2016density} to achieve precise inverse mapping from output to input, thus ensuring information integrity, it is employed as a lossless feature extraction block to further extract detailed features $D_P$ and $D_M$ from the shallow features of both modalities.
The detail feature extraction branch initially divides the shallow features $F_i \in \mathbb{R}^{C \times H \times W}$ along the channel dimension into two parts, $F_{i,f} \in \mathbb{R}^{C/2 \times H \times W}$ and $F_{i,l} \in \mathbb{R}^{C/2 \times H \times W}$. It then computes the final detailed features through three affine coupling layers, with the first layer being expressed as:
\begin{equation}
\begin{gathered}
D_{i,f}^1=F_{i,f} \odot \exp \left(I R B\left(F_{i,l}\right)\right) +I R B\left(F_{i,l}\right), \\
D_{i,l}^1=F_{i,l},
\end{gathered}
\end{equation}
where $IRB(\cdot)$ is the invertible residual block used in MobileNetV2~\upcite{sandler2018mobilenetv2}, and $\odot$ denotes the Hadamard product. After three layers of transformation, $D_i$ is obtained by concatenating two sets of detailed features along the channel dimension.
The design of the independent Detailed Encoder prevents the direct transfer of modality-specific textures. The shared-weight encoders could force different modalities into a common latent space, potentially causing the high-frequency texture of the MRI to overwrite the lower-resolution PET signal. Our independent parameterization ($f_{\theta_p} \neq f_{\theta_m}$) ensures that the detailed features $D_P$ and $D_M$ retain their distinct modality characteristics. This allows the subsequent Cross-Modality Aggregation module to selectively filter inconsistent anatomical edges (pseudo-structures) that appear in the MR image but do not correspond to metabolic activity in the PET image.

\subsubsection{Cross-Modality Aggregation (CMA) Module}
To integrate the high-level features extracted by modality-specific modules, we input the global and detailed features from both the target and auxiliary modalities into a cross-modality aggregation module shown in Fig.~\ref{fig:framework}(c). 
To enhance the deep connections between PET and MR images, we employ a dual-fusion strategy. Specifically, we adopt a global fusion layer and a detail fusion layer to process the corresponding features from both modalities. This process yields the fused global feature $G_{fuse}\in \mathbb{R}^{C \times H \times W}$ and the fused detailed feature $D_{fuse}\in \mathbb{R}^{C \times H \times W}$.
The architecture of the global fusion layer mirrors that of the global feature extraction branch, while the detail fusion layer aligns with the detail feature extraction branch. Subsequently, the decoder network, which includes two Transformer blocks, maps high-level features back to the original input size and outputs the final fused feature $F\in \mathbb{R}^{C \times H \times W}$.

\subsection{Conditional Diffusion Restoration Module}
\label{sec: Conditional Diffusion Restoration Module}
Once the extracted fusion feature has been obtained, we aim to utilize it for SPET image restoration.
Due to the powerful generative capability of diffusion models, \textit{e.g.}, DDPM~\upcite{ho2020denoising}, to learn complex data distributions and sample effectively to produce high-quality images, we adopt it to restore high-quality SPET images.
We inject the fusion feature $F$ (termed as $x_{fusion}$) into the proposed conditional diffusion restoration module as a condition. By incorporating fusion features from LPET and MR images as conditional constraints, the diffusion model is able to generate complex details in SPET images more accurately.

\subsubsection{Forward Process} 
Assuming the target data $x_{0} \sim q\left(x_{0}\right)$, the forward diffusion process can be considered as a Markov chain. By incrementally adding Gaussian noise to the target data as follow, a sequence of latent noise variables $x_1,x_2,...x_t$ can be generated:
\begin{equation}\label{eq1}
q\left(x_{1}, . ., x_{T} \mid x_{0}\right)=\prod_{t=1}^{T} q\left(x_{t} \mid x_{t-1}\right), \forall t \in\{1, \ldots, T\}
\end{equation}
where $q\left(\boldsymbol{x}_{t} \mid \boldsymbol{x}_{t-1}\right)$ represents a single-step Markov process:
\begin{equation}\label{eq2}
q\left(x_{t} \mid x_{t-1}\right)=\mathcal{N}\left(x_{t} ; \sqrt{1-\beta_{t}} \cdot x_{t-1}, \beta_{t} \cdot I\right),
\end{equation}
where $t$ represents the number of diffusion steps, $\left\{\beta_{t}\right\}_{t=1}^{T}$ denotes the variance schedule of Gaussian noise across these steps, with $0<\beta_1<\beta_2<...<\beta_t<1$. This schedule is used to control the amount of noise added at each step. For a predefined variance schedule, when the number of diffusion steps $T$ is sufficiently large, the resulting distribution $q\left(x_{T}\right) \sim \mathcal{N}\left(x_{T} ; 0, I\right)$ becomes stationary.

As defined on a Markov chain, a key characteristic of the diffusion process is the ability to directly sample data $x_t$ at any time step $t$ based on the original data $x_0$. By defining $\alpha_{t}=1-\beta_{t}$ and $\bar{\alpha}_{t}=\prod_{i=0}^{t} \alpha_{i}$, the sampling process can be expressed as:
\begin{equation}
\begin{gathered}
q\left(x_t \mid x_0\right)=\mathcal{N}\left(x_t ; \sqrt{\bar{\alpha}_t} x_0,\left(1-\bar{\alpha}_t\right) \cdot \mathbf{I}\right), \\
x_t=\sqrt{\bar{\alpha}_t} x_0+\sqrt{1-\overline{\alpha_t}} \epsilon,
\end{gathered}
\end{equation}
where $\epsilon \sim \mathcal{N}(\mathbf{0}, \mathbf{I})$.

\subsubsection{Reverse Process}
The reverse process is the inverse of the forward process, which generates target data by gradually denoising noisy samples. To enhance the restoration of SPET images using an auxiliary MR image, we integrate fusion features as additional conditions into the inverse process. 
As shown in Fig.~\ref{fig:framework}, alongside the noisy sample $x_{t}$ and time step $t$ are simultaneously input into the denoising attention UNet at each step. 
To fully leverage the information from both the original LPET $x_{P}$ and the fusion feature $x_{fusion}$ with the MR guided, we also supply these components at each step to guide the denoising process.
It alters the noise prediction model from $\epsilon_{\theta}\left(x_{t}, t\right)$ to $\epsilon_{\theta}\left(x_{t}, t, x_{P},x_{fusion}\right)$, adding two inputs as conditions.
The fusion feature $x_{fusion}$ (denoted as $F$ in Fig.~\ref{fig:framework}(c)), the LPET image $x_P$, and the noisy state $x_t$ are concatenated along the channel dimension to form the input to the first layer of the diffusion U-Net.
For an input sample $x_t$ with $t$ steps noise added, the objective of the noise prediction model is to minimize the distance between the predicted noise and actual added noise. 
Then, with the trained noise prediction model $\epsilon_{\hat\theta}\left(x_{t}, t, x_{P},x_{fusion}\right)$, each refinement step in the reverse process is:

\begin{small}
\begin{align}\label{eq13}
x_{t-1}=\frac{1}{\sqrt{\alpha_{t}}}\left[x_{t}-\frac{\beta_{t}}{\sqrt{1-\bar{\alpha}_{t}}} \epsilon_{\hat{\theta}}\left(x_{t}, t, x_{P},x_{fusion}\right)\right]&+\sigma_{t} z,
\end{align}
\end{small}
where $z \sim \mathcal{N}(\mathbf{0}, \mathbf{I})$.

\subsection{Two-stage Supervise-assisted Learning}
\label{sec: Two-stage Supervise-assisted Learning Strategy}
Acquiring PET images poses significant challenges, often resulting in small clinical PET datasets. Variability in scanning times, tracer types, injection doses, and methods customized for individual patient needs contribute to the generation of out-of-distribution (OOD) data. Deep learning approaches that train and test under uniform conditions excel on test sets mirroring the training data distribution but falter when confronted with real-world clinical variations. These methods struggle with low-dose PET images and simulations, leading to notable performance degradation when encountering diverse OOD data.

To address these challenges, we introduce a two-stage supervise-assisted learning strategy with the loss in Eq. (\ref{eq14}). (\textbf{i}) In the first stage, \textit{i.e.}, \textbf{external stage}, the model MFdiff is trained on a large-scale simulated phantom dataset. We leverage the simulated data to learn multimodal feature fusion and high-quality PET image generation, thereby avoiding the high costs associated with extensive clinical data usage. Using the diffusion model, we supervise the learning of noise at each prediction step. The loss function can be expressed as:
\begin{equation}\label{eq14}
\mathcal{L}=E_{t, x_{0}, \epsilon}\left[\left\|\epsilon-\epsilon_{\theta}\left(x_{t}, t, x_{P},x_{fusion}\right)\right\|^{2}\right].
\end{equation}
Thus, we derive generalized priors from the large-scale simulated dataset and initialize the model parameters.
(\textbf{ii}) In the second stage, \textit{i.e.}, \textbf{internal stage}, we further train the initialized model on a small set of clinically realistic in-vivo data to learn specific priors for in-vivo OOD data. The supervised training approach in this stage aligns with Eq. (\ref{eq14}), allowing us to further discern specific differences through OOD data, which enhances its adaptability across various scenarios with minimal practical overhead. 

\textit{Learning details.} During the training in the external stage, the model processes data from the simulated dataset, comprising 1000 samples from 20 brain phantoms, distributed across training, validation, and test sets at ratios of 0.8, 0.15, and 0.05, respectively. In the internal stage, we conduct training on a small cohort of clinical data, encompassing varied scanning times and injection dosages.

\section{Experiments}
In this section, we first introduce our experimental settings, including the comparison methods, evaluation metrics, and implementation details. We then introduce the simulated phantom dataset and three in-vivo out-of-distribution datasets used for experiments.

\subsection{Experimental Settings}

\subsubsection{Comparison Methods}
In order to verify the superiority of our method, the following 6 state-of-the-art image restoration or denoising methods are implemented for comparison, including M-UNet~\upcite{chen2019ultra}, FBSEM~\upcite{mehranian2020model}, EA-GAN~\upcite{yu2019ea}, Hi-Net~\upcite{zhou2020hi}, CNCL~\upcite{geng2021content}, and CSRD~\upcite{yoon2024volumetric}. 
Following the work~\upcite{luo2022adaptive}, we implement the networks that are originally set to 2.5D or 3D in the comparison methods as 2D networks (changing the input dimension to match our slice-based approach). We also add MR images as additional input, ensuring all models have access to the same multimodal information.
For all comparison methods, we adopt the hyperparameters (\textit{e.g.}, loss weighting) mentioned in the respective papers and train the models with the same settings (\textit{e.g.}, iterations, learning rates, and batch sizes) as ours.

\subsubsection{Evaluation Metrics}
To evaluate and compare the performance of different methods comprehensively, three typical evaluation metrics are used, including peak signal to noise ratio (PSNR)~\upcite{huynh2008psnr}, structural similarity (SSIM)~\upcite{wang2004ssim} and normalized mean square error (NMSE)~\upcite{wang2009mse}. The larger PSNR and SSIM values the better the restored images are, while the smaller values of NMSE indicate better restoration. 

\subsubsection{Implementation Details}
The proposed method is implemented through the Pytorch framework, trained with an NVIDIA GeForce RTX 3090 GPU. The denoising network used in the experiment originated from the UNet architecture with time embedding modules and self-attention layers. During network training, the size of input LPET and auxiliary MR image are both $192\times192$, the batch size is 4, and the number of time step is 1000. 
If not mentioned otherwise, the same hyperparameters as those in the work~\upcite{ho2020denoising} are used.
We train the model for 50,000 iterations in the external stage and 20,000 iterations in the internal stage, using the AdamW optimizer~\upcite{kinga2015method} with learning rate of 0.0001.

\subsection{Simulated Phantom Data}
We leverage the BrainWeb~\upcite{cocosco1997brainweb, collins1998design}, which has 20 3D brain phantoms to construct the simulated dataset. For PET data, by modeling the acquisition geometry of the Siemens Biograph mMR, the simulated PET data has a voxel size of $2.08\times2.08\times2.03 mm^3$ and an image resolution of $344\times344\times127$. For each 3D brain phantom, we select 10 noncontinuous slices in the middle and perform data augmentation. Each slice is rotated along the axial direction by 5 random angles in the range 0-15, resulting in 1000 data slices. For each data slice, a high-definition SD sinogram and a low-definition LD sinogram are generated. For SD sinograms, we adopt $1\times10^{10}$ counts and point spread function modeling using 2.5~mm full-width at half-maximum (FWHM) Gaussian kernels. The OSEM algorithm~\upcite{hudson1994accelerated} is then used to restore SPET images, and the number of iterations and subsets are set to 15 and 10, respectively. Similarly, for LD sinograms, randomly select count levels between $9\times10^{7}$ and $12\times10^{7}$, with 4~mm FWHM Gaussian kernels. To restore SPET images, the OSEM algorithm is used with 10 iterations and 14 subsets.
The split of 1000 slices from 20 brain phantoms is performed at the subject (phantom) level before slicing. This ensures that slices from the same phantom do not appear in both training and testing, preventing subject-level leakage.

\subsection{In-Vivo Data}
To verify and compare the generalization performance of all methods, we adopt in-vivo out-of-distribution (OOD) datasets in three different imaging scenarios, including (1) variation in imaging time, (2) variation in radiotracer injection dose, and (3) variation in radiotracer administration protocol. The following details the acquisition, data preprocessing, and data augmentation of OOD datasets.

\subsubsection{OOD Data with Variation in Imaging Time}
For the OOD data-1, we utilize an open-access dataset\upcite{jamadar2020simultaneous} acquired from a Siemens (Erlangen) Biograph 3-Tesla molecular MR (mMR) scanner at Monash Biomedical Imaging in Melbourne, Australia. Participants include 27 healthy volunteers aged 18-23 years. They are perfused with ${}^{18}$F-Flurodeoxyglucose (${}^{18}$F-[FDG]), with average dose 233MBq and at a rate of 36 mL/hour. Scans take 95 minutes. Non-functional MR scans are performed 30 minutes after perfusion, corresponding to T1-MR images with FOV = $256\times256 \ mm^2$, voxel size = $1\times1\times1 \ mm^3$, 176 slices. Afterwards, 3600 s of list-mode PET data for each subject starting at the 30-minute time point are merged into 225 3D sinogram frames, with a single PET frame originating from a scan of 16 s. The 3D PET image is restored through the OSEM algorithm with 3 iterations and 21 subsets, and then a 5 mm FWHM Gaussian post-processing filter is applied to each 3D volume, and all 3D PET volumes are spliced into a 4D volume with a size of $344\times344\times127\times225$ in time series.

The SPM12 toolbox\footnote{https://www.fil.ion.ucl.ac.uk/spm/software/spm12/} is used to register the PET images to the T1-MR images, and then all image pixel intensities are normalized to range of $[0, 1]$. For each subject, we select 10 spaced slices in the third dimension from 130 to 150, and construct SPET images using a 30-minute scan. By reducing the acquisition time, LPET images with two settings of 2min scan and single 16s scan are constructed. For each slice, the same data augmentation method as simulated data is used.

\subsubsection{OOD Data with Variation in Radiotracer Injection Dose}
The MR/PET data acquisition protocol for OOD data-2 is the same as that for OOD data-1, \textit{i.e.}, using the same mMR scanner, radiotracer, and imaging settings, \textit{etc.}, but with different injection doses. For the LD-PET data, the average injected dose is 225 MBq and average PSNR is 23.74 dB, and for the ultra low-dose PET (uLD-PET) data, the average injected dose is 167 MBq and average PSNR is 21.35 dB. The data preprocessing and data augmentation methods of OOD data-2 are consistent with those used for OOD data-1.

\subsubsection{OOD Data with Variation in Radiotracer Administration Protocol}
The data acquisition protocol for OOD data-3 is the same as that for OOD data-1 and data-2. However, the radiotracer administration protocol varies, incorporating three forms: bolus, constant infusion, and hybrid 50\% bolus-50\% infusion. For all protocols, the bolus or infusion is administered concurrently with the start of the PET scan, and the infusion rate is set as 36 mL/hour. For both infusion and hybrid bolus-infusion administration, the infusion is discontinued at 55 minutes. 
For additional details on data acquisition and procedures, refer to work~\upcite{jamadar2022monash}.
We select 300 pairs of typical data (6 subjects) for all in-vivo OOD data tested in our experiments, including 50 pairs (1 subject) for OOD data-1, 100 pairs (2 subjects) for OOD data-2, and 150 pairs (3 subjects) for OOD data-3. For each type of data, we adopt 250 pairs of data (5 subjects) for internal stage training and validation (10\%).

\section{Results}
\subsection{Simulation Study}

Table~\ref{tab:phantom} presents the quantitative results on the phantom brain dataset in the form of $u\pm v$, where $u$ and $v$ represent the average value and the standard deviation value of each evaluation metric, respectively. From Table~\ref{tab:phantom}, it can be observed that our MFdiff outperforms all compared methods in PSNR, SSIM, and NMSE. This indicates that MFdiff can effectively utilize information from different modalities to achieve high-quality SPET image restoration.

\begin{table}[t]
\centering
\small
\caption{Quantitative results (Mean$\pm$Standard Deviation) associated with state-of-the-art PET image restoration methods in terms of PSNR, SSIM and NMSE on Phantom brain dataset. \textbf{Bold} and \underline{Underline} indicate the best and second-best method for each quantitative evaluation, respectively.}
\setlength\tabcolsep{1.5mm}
\begin{tabular}{l c c c c c c c}
\toprule
Method & PSNR (dB) $\uparrow$ & SSIM $\uparrow$ & NMSE $\downarrow$ \\
\midrule
LPET & 21.13 $\pm$ 2.71 & 0.839 $\pm$ 0.041 & 0.176 $\pm$ 0.055 \\
M-UNet~\upcite{chen2019ultra} & 31.35 $\pm$ 2.59 & 0.981 $\pm$ 0.007 & 0.018 $\pm$ 0.009 \\
FBSEM~\upcite{mehranian2020model} & 34.93 $\pm$ 2.03 & 0.959 $\pm$ 0.009 & 0.039 $\pm$ 0.021 \\
EA-GAN~\upcite{yu2019ea} & 32.09 $\pm$ 2.10 & 0.962 $\pm$ 0.036 & 0.015 $\pm$ 0.009 \\
Hi-Net~\upcite{zhou2020hi} & 32.65 $\pm$ 2.15 & 0.981 $\pm$ 0.008 & 0.013 $\pm$ 0.006 \\
CNCL~\upcite{geng2021content} & 33.32 $\pm$ 1.78 & 0.984 $\pm$ 0.004 & 0.016 $\pm$ 0.009 \\
CSRD~\upcite{yoon2024volumetric} & \underline{37.28 $\pm$ 1.94} & \underline{0.990 $\pm$ 0.006} & \underline{0.009 $\pm$0.005} \\
MFdiff (Ours) & \textbf{37.93 $\pm$ 1.85} & \textbf{0.991 $\pm$ 0.008} & \textbf{0.008 $\pm$ 0.004} \\
\bottomrule
\end{tabular}
\label{tab:phantom}
\end{table}

\begin{table*}[t]
\caption{Quantitative results (Mean$\pm$Standard Deviation) associated with state-of-the-art PET image restoration methods in terms of PSNR, SSIM and NMSE on OOD Data-1.
}
\centering
\small
\renewcommand{\arraystretch}{1.1}
\tabcolsep=0.02\linewidth
\begin{tabular}{c|ccc|ccc}
\hline
\multirow{2}{*}{Method} & \multicolumn{3}{c|}{\textbf{16s scanning-PET}} & \multicolumn{3}{c}{\textbf{2min scanning-PET}} \\
& PSNR (dB) $\uparrow$ & SSIM $\uparrow$ & NMSE $\downarrow$ & PSNR (dB) $\uparrow$ & SSIM $\uparrow$ & NMSE $\downarrow$ \\
\hline
LPET & 25.01 $\pm$ 1.21 & 0.928 $\pm$ 0.004 & 0.037 $\pm$ 0.009  & 34.65 $\pm$ 1.24 & 0.987 $\pm$ 0.001 & 0.004 $\pm$ 0.002 \\
M-UNet~\upcite{chen2019ultra} & 27.04 $\pm$ 1.86 & 0.949 $\pm$ 0.004 & 0.013 $\pm$ 0.006  & 32.57 $\pm$ 0.92 & 0.977 $\pm$ 0.001 & \underline{0.003 $\pm$ 0.001} \\
EA-GAN~\upcite{yu2019ea} & 27.16 $\pm$ 1.68 & 0.970 $\pm$ 0.003 & 0.019 $\pm$ 0.007 & 33.83 $\pm$ 1.14 & 0.949 $\pm$ 0.008 & 0.005 $\pm$ 0.001 \\
Hi-Net~\upcite{zhou2020hi} & 27.95 $\pm$ 1.86 & 0.970 $\pm$ 0.002 & 0.020 $\pm$ 0.009  & 30.02 $\pm$ 1.81 & 0.985 $\pm$ 0.001 & 0.013 $\pm$ 0.006 \\
CNCL~\upcite{geng2021content} & 28.77 $\pm$ 1.54 & \underline{0.974 $\pm$ 0.002} & 0.014 $\pm$ 0.005 & 34.36 $\pm$ 1.24 & 0.976 $\pm$ 0.007 & 0.004 $\pm$ 0.001 \\
CSRD~\upcite{yoon2024volumetric} & \underline{29.90 $\pm$ 1.39} & 0.973 $\pm$ 0.014 & \underline{0.012 $\pm$ 0.004} & \underline{36.05 $\pm$ 0.98} & \underline{0.987 $\pm$ 0.005} & \textbf{0.002 $\pm$ 0.001}\\
MFdiff (Ours) & \textbf{30.67 $\pm$ 1.37} & \textbf{0.974 $\pm$ 0.011} & \textbf{0.010 $\pm$ 0.003} & \textbf{36.97 $\pm$ 0.98} & \textbf{0.989 $\pm$ 0.005} & \textbf{0.002 $\pm$ 0.001} \\
\hline
\end{tabular}
\label{tab:ood1}
\end{table*}

\begin{figure*}[t]
\centering
\includegraphics[width=1\textwidth]{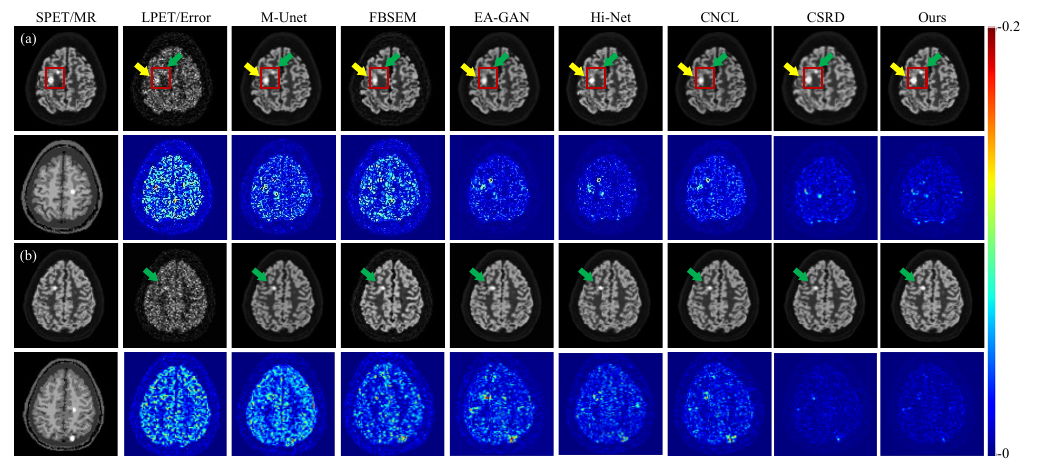}
\caption{Visual comparison results of the restored PET along with error maps by different methods on phantom dataset.}
\label{fig:phantom}
\end{figure*}

\begin{table*}[t]
\caption{Quantitative results (Mean$\pm$Standard Deviation) associated with state-of-the-art PET image restoration methods in terms of PSNR, SSIM and NMSE on OOD Data-2. }
\centering
\small
\renewcommand{\arraystretch}{1.1}
\tabcolsep=0.02\linewidth
\begin{tabular}{c|ccc|ccc}
\hline
\multirow{2}{*}{Method} & \multicolumn{3}{c|}{\textbf{LD-PET}} & \multicolumn{3}{c}{\textbf{uLD-PET
}} \\
& PSNR (dB) $\uparrow$ & SSIM $\uparrow$ & NMSE $\downarrow$ & PSNR (dB) $\uparrow$ & SSIM $\uparrow$ & NMSE $\downarrow$ \\
\hline
LPET & 21.35 $\pm$ 1.16 & 0.880 $\pm$ 0.008 & 0.084 $\pm$ 0.021 & 23.74 $\pm$ 0.92 & 0.905 $\pm$ 0.008 & 0.046 $\pm$ 0.009 \\
M-UNet~\upcite{chen2019ultra} & 27.89 $\pm$ 0.65 & 0.945 $\pm$ 0.003 & 0.009 $\pm$ 0.002 & 26.22 $\pm$ 1.57 & 0.936 $\pm$ 0.010 & 0.013 $\pm$ 0.005 \\
EA-GAN~\upcite{yu2019ea} & 27.33 $\pm$ 0.85 & 0.941 $\pm$ 0.004 & 0.017 $\pm$ 0.003 & 27.88 $\pm$ 1.35 & 0.933 $\pm$ 0.004 & 0.016 $\pm$ 0.005 \\
Hi-Net~\upcite{zhou2020hi} & 24.67 $\pm$ 1.02 & 0.948 $\pm$ 0.012 & 0.038 $\pm$ 0.010 & 27.80 $\pm$ 1.31 & 0.964 $\pm$ 0.003 & 0.020 $\pm$ 0.007 \\
CNCL~\upcite{geng2021content} & 26.35 $\pm$ 1.10 & 0.954 $\pm$ 0.004 & 0.020 $\pm$ 0.004 & 27.34 $\pm$ 1.54 & 0.956 $\pm$ 0.004 & 0.018 $\pm$ 0.006 \\
CSRD~\upcite{yoon2024volumetric} & \underline{28.48 $\pm$ 1.29} & \textbf{0.966 $\pm$ 0.012} & \underline{0.015 $\pm$ 0.004}  & \underline{29.69 $\pm$ 0.99} & \underline{0.965 $\pm$ 0.009} & \textbf{0.009 $\pm$ 0.002} \\
MFdiff (Ours) & \textbf{28.96 $\pm$ 1.25} & \underline{0.965 $\pm$ 0.012} & \textbf{0.014 $\pm$ 0.004} & \textbf{30.37 $\pm$ 1.02} & \textbf{0.972 $\pm$ 0.005} & \underline{0.010 $\pm$ 0.002} \\
\hline
\end{tabular}
\label{tab:ood2}
\end{table*}
%

\begin{table*}[t]
\caption{Quantitative results (Mean$\pm$Standard Deviation) associated with state-of-the-art PET image restoration methods in terms of PSNR, SSIM and NMSE on OOD Data-3. }
\centering
\small
\renewcommand{\arraystretch}{1.1}
\tabcolsep=0.004\linewidth
\scalebox{0.9}{
\begin{tabular}{c|ccc|ccc|ccc}
\hline
\multirow{2}{*}{Method} & \multicolumn{3}{c|}{\textbf{Bolus (B)}} & \multicolumn{3}{c|}{\textbf{Infusion (I)}} & \multicolumn{3}{c}{\textbf{Bolus + Infusion (B/I)}} \\
& PSNR (dB) $\uparrow$ & SSIM $\uparrow$ & NMSE $\downarrow$ & PSNR (dB) $\uparrow$ & SSIM $\uparrow$ & NMSE $\downarrow$ & PSNR (dB) $\uparrow$ & SSIM $\uparrow$ & NMSE $\downarrow$ \\
\hline
LPET& 19.92 $\pm$ 0.56 & 0.867 $\pm$ 0.004 & 0.097 $\pm$ 0.012 & 20.03 $\pm$ 1.35 & 0.886 $\pm$ 0.008 & 0.119 $\pm$ 0.030 & 24.45 $\pm$ 2.44 & 0.940 $\pm$ 0.011 & 0.092 $\pm$ 0.036\\
M-UNet~\upcite{chen2019ultra}& 27.36 $\pm$ 1.05 & 0.892 $\pm$ 0.039 & 0.018 $\pm$ 0.005 & \underline{30.03 $\pm$ 1.17} & 0.893 $\pm$ 0.053 & \underline{0.012 $\pm$ 0.003} & \underline{30.33 $\pm$ 2.04} & 0.899 $\pm$ 0.077 & 0.028 $\pm$ 0.027\\
EA-GAN~\upcite{yu2019ea} & 27.47 $\pm$ 1.30 & 0.878 $\pm$ 0.053 & 0.018 $\pm$ 0.006 & 29.88 $\pm$ 0.68 & 0.853 $\pm$ 0.077 & \underline{0.012 $\pm$ 0.003} & 28.34 $\pm$ 2.08 & 0.916 $\pm$ 0.034 & \underline{0.017 $\pm$ 0.009} \\
Hi-Net~\upcite{zhou2020hi}& 27.32 $\pm$ 1.26 & 0.929 $\pm$ 0.036 & 0.019 $\pm$ 0.010 & 29.76 $\pm$ 0.82 & 0.950 $\pm$ 0.014 & 0.013 $\pm$ 0.003 & 26.45 $\pm$ 3.27 & 0.901 $\pm$ 0.057 & 0.127 $\pm$ 0.201\\
CNCL~\upcite{geng2021content}& \underline{27.89 $\pm$ 0.80} & 0.898 $\pm$ 0.027 & \underline{0.015 $\pm$ 0.003} & 29.32 $\pm$ 1.17 & 0.888 $\pm$ 0.040 & 0.014 $\pm$ 0.003 & 27.35 $\pm$ 1.68 & 0.932 $\pm$ 0.022 & 0.019 $\pm$ 0.006\\
CSRD~\upcite{yoon2024volumetric} & 27.88 $\pm$ 0.91 & \underline{0.939 $\pm$ 0.017} & 0.015 $\pm$ 0.004 & 28.86 $\pm$ 1.07 & \underline{0.960 $\pm$ 0.014} & 0.015 $\pm$ 0.004 & 29.13 $\pm$ 1.07 & \underline{0.941 $\pm$ 0.014} & \textbf{0.017 $\pm$ 0.005}\\
MFdiff (Ours)& \textbf{28.23 $\pm$ 0.71} & \textbf{0.940 $\pm$ 0.009} & \textbf{0.014 $\pm$ 0.003} & \textbf{30.41 $\pm$ 0.93} & \textbf{0.967 $\pm$ 0.009} & \textbf{0.010 $\pm$ 0.002} & \textbf{31.92 $\pm$ 2.35} & \textbf{0.949 $\pm$ 0.032} & 0.019 $\pm$ 0.016\\
\hline
\end{tabular}}
\label{tab:ood3}
\end{table*}

In Fig.~\ref{fig:phantom}, we select two representative slices with lesions for visual comparison, and display the restored SPET images along with error maps by different methods shown in Figs. \ref{fig:phantom}(a-b).
Notably, the upper circular lesion (indicated by the green arrow in Fig.~\ref{fig:phantom}(a)) represents a critical test case for MR-induced artifact suppression. In this region, the anatomical MRI (first column) displays healthy tissue structure, which contradicts the metabolic lesion present in the ground-truth SPET. Comparison methods like M-UNet and FBSEM rely on direct concatenation or naive fusion, which inadvertently transfers healthy MRI textures into the PET image. This interference causes the blurring or partial removal of lesions, effectively acting as a false negative. In contrast, MFdiff correctly preserves the shape and intensity of lesions. This validates that the IML module effectively disentangles the modalities, utilizing MRI for global anatomical guidance without hallucinating MRI-specific structures where they are metabolically inconsistent.

\begin{figure}[t]
\centering
\includegraphics[width=1.05\columnwidth]{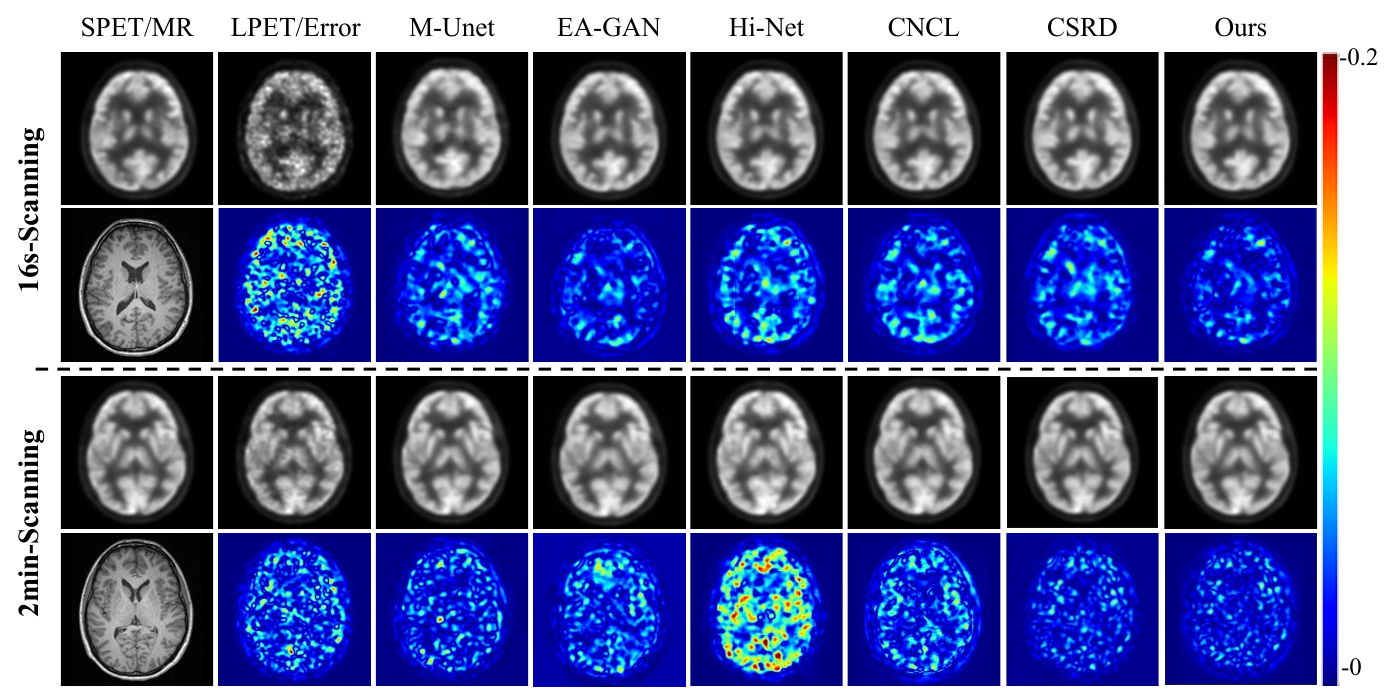}
\caption{Visual comparison along with error maps of different methods on OOD data-1 with variation in imaging time.}
\label{fig:ood1}
\end{figure}

\begin{figure}[t]
\centering
\includegraphics[width=1.05\columnwidth]{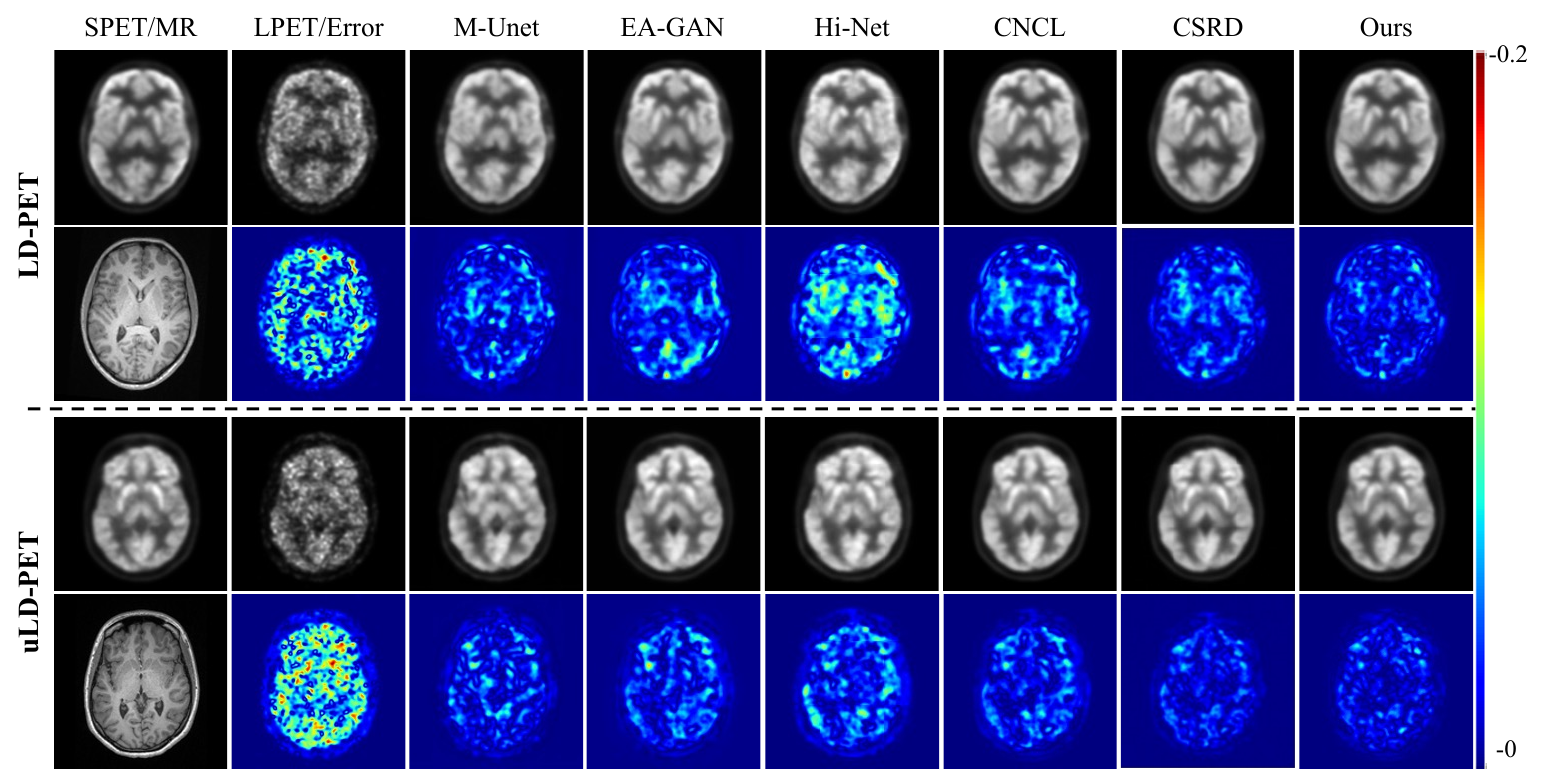}
\caption{Visual comparison along with error maps of different methods on OOD data-2 with variation in radiotracer injection dose.}
\label{fig:ood2}
\end{figure}

\begin{figure}[t]
\centering
\includegraphics[width=1.05\columnwidth]{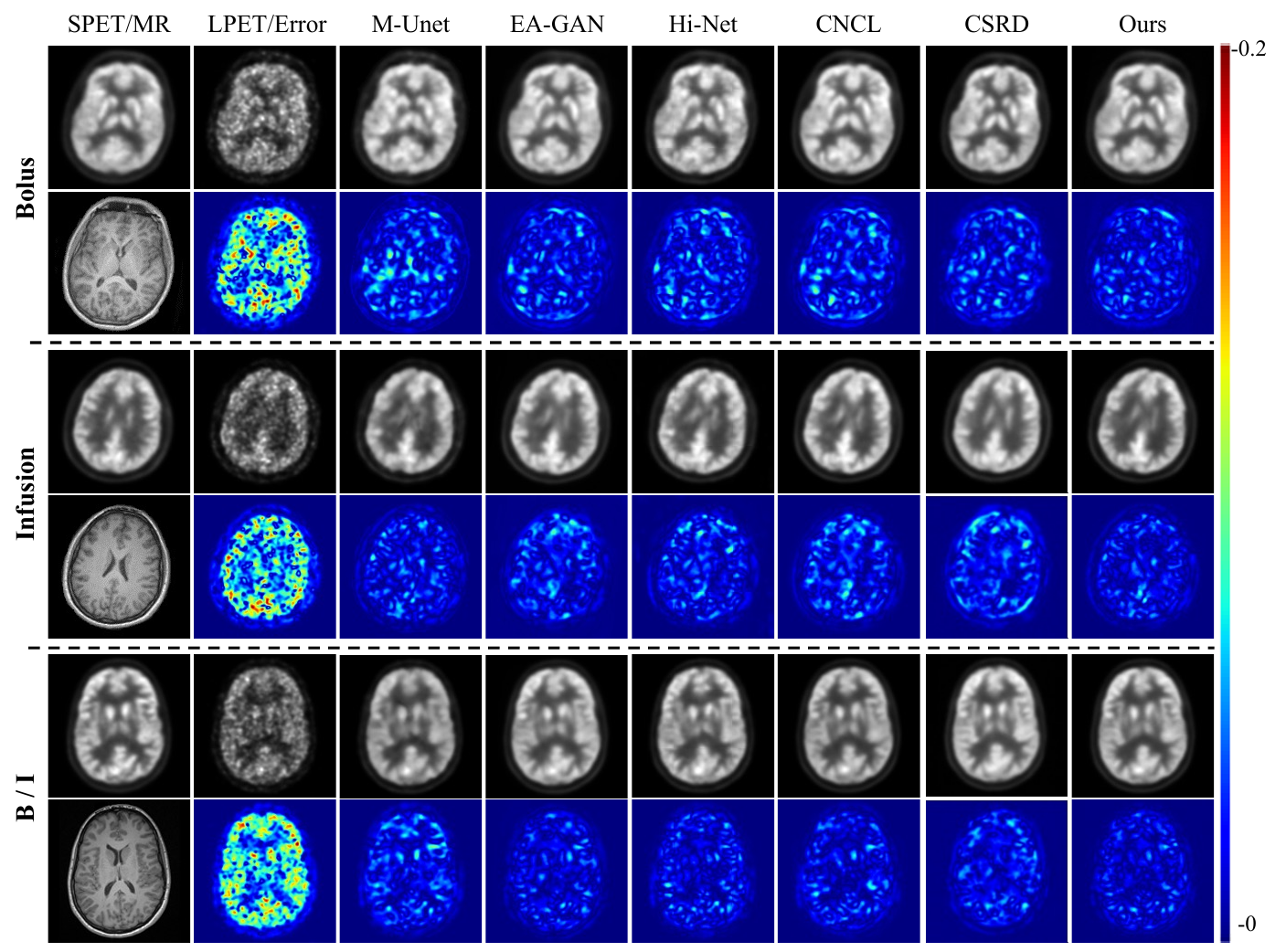}
\caption{Visual comparison along with error maps of different methods on OOD data-3 with variation in radiotracer administration protocols.}
\label{fig:ood3}
\end{figure}

\begin{table*}[ht]
\centering
\small
\renewcommand{\arraystretch}{1.1}
\tabcolsep=0.012\linewidth
\caption{Quantitative results (Mean$\pm$Standard Deviation) of the ablation
study over the Phantom sataset.}
{
\begin{tabular}{l c c c c c c c c c c c c}
\toprule
Method & Multi-input & Global Feature & Detailed Feature & CDDPM & PSNR (dB) $\uparrow$ & SSIM $\uparrow$ & NMSE $\downarrow$ \\
\midrule
UNet & - & - & - & - & 26.39 $\pm$ 1.39 & 0.924 $\pm$ 0.009 & 0.050 $\pm$ 0.009 \\
MUNet & \checkmark & - & - & - &  33.35 $\pm$ 2.59 & 0.981 $\pm$ 0.007 & 0.018 $\pm$ 0.009 \\
GMUNet & \checkmark & \checkmark & - & - & 36.39 $\pm$ 2.59 & 0.985 $\pm$ 0.006 & 0.009 $\pm$ 0.005 \\
DMUNet & \checkmark & - & \checkmark & - & 35.93 $\pm$ 1.56 & 0.987 $\pm$ 0.005 & 0.009 $\pm$ 0.004 \\
CMUNet &\checkmark &\checkmark &\checkmark & - & 36.51 $\pm$ 2.59 & 0.984 $\pm$ 0.011 & 0.009 $\pm$ 0.005 \\
MFdiff (Ours) &\checkmark &\checkmark &\checkmark & \checkmark & \textbf{37.93 $\pm$ 1.85} & \textbf{0.991 $\pm$ 0.008} & \textbf{0.008 $\pm$ 0.004} \\
\bottomrule
\end{tabular}}
\label{tab:ablation}
\end{table*}

\subsection{Clinical Study}
Clinical real data often exhibit significant distributional differences due to varied operational practices, so the ability of models to adapt to such distributionally inconsistent data, \textit{i.e.}, generalization, is an important basis for measuring the value of its practical application. We conduct experiments on in-vivo OOD data generated in three different situations with the model trained on simulated datasets and further on a modest amount of clinical real data. In clinical study, FBSEM is excluded from comparison because it requires sinogram domain data, which is not present in these datasets.

Tables~\ref{tab:ood1}-\ref{tab:ood3} illustrate the quantitative restoration results across changes in imaging time, radiotracer injection dose, and radiotracer administration protocol on three in-vivo OOD datasets. 
The results demonstrate that our MFdiff consistently achieves higher quantitative scores compared to other methods in all three scenarios.
Specifically, for the 2-minute scan setting in Table~\ref{tab:ood1}, M-UNet, EA-GAN, Hi-Net, and CNCL enhance the visual quality of restoration compared to low-dose PET images, as depicted in Fig.~\ref{fig:ood1}. Despite these visual enhancements, the quantitative performance indices for these methods show some deterioration compared to LPET images. 
Notably, the average PSNR for MFdiff method under 2-minute setting improves from 34.65 dB to 36.97 dB by 0.92 dB over the second-best method (CSRD), demonstrating the adaptability.
Figs.~\ref{fig:ood1}-\ref{fig:ood3} show the qualitative results of all methods on three in-vivo OOD datasets, respectively. 
The results illustrate that MFdiff predicts missing information in LPET images more accurately, resulting in restored images with clearer edges, structures, and rich texture details. From the error maps, it can be observed that MFdiff achieves the smallest restoration errors across all in-vivo OOD data scenarios, indicating that the restored images closely match those of LPET images.

\begin{table}[t]
\centering
\small
\renewcommand{\arraystretch}{1.1}
\caption{Quantitative Results of the Ablation
Study of two-stage supervise-assisted learning strategy on OOD data-1. External}
\setlength\tabcolsep{1mm}
\scalebox{0.85}{
\begin{tabular}{c c c c c c c c c}
\toprule
Settings & Learning & PSNR (dB) $\uparrow$ & SSIM $\uparrow$ & NMSE $\downarrow$ \\
\midrule
16s & External & 21.91 $\pm$ 3.05 & 0.718 $\pm$ 0.208 & 0.107 $\pm$ 0.140\\
Scanning & Internal & 29.81 $\pm$ 1.74 & \textbf{0.975 $\pm$ 0.008} & 0.013 $\pm$ 0.005 \\
PET & External+Internal & \textbf{30.67 $\pm$ 1.37} & 0.974 $\pm$ 0.011 & \textbf{0.010 $\pm$ 0.003} \\ \midrule
2min & External & 31.35 $\pm$ 2.59 & 0.981 $\pm$ 0.007 & 0.018 $\pm$ 0.009\\
Scanning & Internal & 35.63 $\pm$ 0.95 & 0.984 $\pm$ 0.009 & 0.003 $\pm$ 0.001 \\
PET & External+Internal & \textbf{36.97 $\pm$ 0.98} & \textbf{0.989 $\pm$ 0.005} & \textbf{0.002 $\pm$ 0.001} \\
\bottomrule
\end{tabular}}
\label{tab:ablation_strategy}
\end{table}

\begin{table}[t]
\centering
\small
\renewcommand{\arraystretch}{1.1}
\caption{Comparison Results of the single-stage (mixed training) and our two-stage learning strategy on OOD data-1 (2 min scanning PET).}
\setlength\tabcolsep{1mm}
\scalebox{0.85}{
\begin{tabular}{c c c c c c c c c}
\toprule
Settings & Learning & PSNR (dB) $\uparrow$ & SSIM $\uparrow$ & NMSE $\downarrow$ \\
\midrule
\multirow{2}{*}{CSRD~\upcite{yoon2024volumetric}} & Single-Stage & 35.17 $\pm$ 0.90 & 0.984 $\pm$ 0.002 & 0.005 $\pm$ 0.002\\
 & Two-Stage & 36.05 $\pm$ 0.98 & 0.987 $\pm$ 0.005 & 0.002 $\pm$ 0.001 \\ \midrule
\multirow{2}{*}{MFdiff (Ours)} & Single-Stage & 35.89 $\pm$ 0.85 & 0.983 $\pm$ 0.002 & 0.003 $\pm$ 0.001\\
 & Two-stage & \textbf{36.97 $\pm$ 0.98} & \textbf{0.989 $\pm$ 0.005} & \textbf{0.002 $\pm$ 0.001} \\
\bottomrule
\end{tabular}}
\label{tab:ablation_strategy_stage}
\end{table}

\subsection{Ablation Study}
The proposed MFdiff comprises several key components. We conduct the following ablation experiments to evaluate the contribution of each module separately: (1) UNet: Using UNet~\upcite{ronneberger2015u} as the restoration network with single LPET image as input; (2) MUNet: Introducing multi-modalily input and employing direct concatenation as the fusion strategy; (3-5) Replacing naive fusion strategy with the multi-modality feature fusion module consisting only of global feature extraction branch, only of detail feature extraction branch and the complete dual-branch, denoted as GMUNet, DMUNet, and CMUNet, respectively; (6) Employing the proposed MFdiff with conditional denoising diffusion probabilistic model (CDDPM) as the restoration module for SPET restoration.
Table~\ref{tab:ablation} shows the quantitative evaluation results for the restored SPET images when comparing our full model with its various ablated versions. 

\textbf{(i) Multi-modality input.} Compared to UNet, MUNet adds MR image as additional input to guide the SPET restoration. The naive concatenation fusion strategy alone increases PSNR by 6.96 dB, SSIM by 0.057, and decreases NMSE by 0.032, indicating that the effectiveness of utilizing multi-modality inputs for this task.
\textbf{(ii) Feature extraction module.} To effectively extract and aggregate complementary information from two modalities, we propose a multi-modality feature fusion module. To investigate the effect of this module as a whole and its crucial component, we compare four ablated versions: MUNet, GMUNet, DMUNet, and CMUNet. To ensure the network parameters of different versions are similar, GMUNet replaces the detail feature extraction branch with a global feature extraction branch, and similarly, DMUNet replaces the global branch with a detail branch. The results in Table~\ref{tab:ablation} demonstrate that incorporating different versions of the fusion module leads to enhancing PET restoration compared to the concatenation. Moreover, CMUNet outperforms the others, indicating that both global and detail branches are beneficial for enhancing the restoration quality, and they mutually reinforce each other.
\textbf{(iii) Diffusion module.} We replace the basic UNet with the diffusion module, CDDPM to compose our MFdiff. The results indicate that the CDDPM leads to increases of PSNR from 36.51 dB to 37.93 dB, demonstrating that CDDPM outperforms the commonly used UNet architecture for medical imaging, and exhibiting a stronger ability for restoring high-quality images.
\textbf{(iv) Learning strategy.} As shown in Table~\ref{tab:ablation_strategy}, it can be seen that a model only trained with simulation data (external stage) performs poorly on OOD data-1 under various experimental conditions. Using our proposed two-stage supervised-assisted learning strategy can improve the performance, compared to training only with external or internal data, and enhance its ability to generalize to specific out-of-distribution (OOD) data.
\textbf{(v) Learning Stage.} As shown in Table~\ref{tab:ablation_strategy_stage}, the `Single-Stage (Mixed Training)' strategy based on MFdiff achieves a PSNR of 35.89 dB. It is superior to `Internal Only' (35.63 dB) but inferior to our proposed `Two-Stage' strategy (36.97 dB). The performance change is also the same on the CSRD model. Therefore, we can see that the large-scale simulated data allows the model to learn stable generalized priors (noise distribution and basic anatomical fusion) without the interference of the domain gap. The second stage then fine-tunes these priors to the specific in-vivo distribution. In contrast, the `Single-Stage (Mixed Training)' strategy forces the model to simultaneously reconcile the domain shift between simulated physics and real scanner noise. It leads to sub-optimal convergence where the model struggles to prioritize the subtler clinical features against the bulk of simulated data.

\section{Conclusion}
In this paper, we propose a novel supervise-assisted multi-modality fusion diffusion model (MFdiff) to restore high-quality SPET images from LPET and corresponding MR images. 
With multi-modality fusion module, specific and complementary features from different modalities can be better extracted and effectively aggregated to generate fusion feature to facilitate SPET restoration. 
The high-quality SPET image restoration module with the powerful data learning and generation capabilities of diffusion model, utilizes multi-modality fusion feature as additional condition to iteratively generate high-quality SPET images. 
Additionally, we introduce a two-stage supervision-assisted learning strategy. In the first external stage, the model is trained on a simulated phantom dataset to capture generalized priors. Subsequently, in the second internal stage, it is fine-tuned on a limited amount of in vivo data. It effectively enables the learning of both generalized and domain-specific priors for SPET restoration. This approach reduces reliance on large multi-modality in-vivo datasets for the target domain and improves adaptability of the model to varying acquisition constraints.
 
The results on simulated dataset and in-vivo OOD datasets show that MFdiff outperforms existing state-of-the-art PET restoration methods in both qualitative and quantitative metrics. It indicates that our method can restore high-quality SPET images more effectively and has better robustness against reductions in scan time and dosage.
Furthermore, in visual comparisons on the simulated phantom dataset, our method effectively integrates multi-modality information and restores the structure of simulated lesion areas more accurately than competing methods. Our clinical evaluation is limited to healthy volunteers, meaning real pathological uptake was not directly assessed. However, we observe improved structural fidelity in simulated lesions and low error rates in complex healthy anatomy. These results suggest that MFdiff effectively preserves critical details without introducing MR-induced artifacts.

\noindent
\textbf{Limitation.} Our method does have limitations that require further improvement. First, the in-vivo evaluation is conducted on a limited cohort of 6 healthy subjects using a single scanner and a single tracer. While the model demonstrates data efficiency by fine-tuning on a small internal dataset, the generalization to multi-center data, different PET scanners, or different radiotracers remains to be validated. In the future, we plan to construct additional datasets with diverse settings to explore their impact on the transferability of the pre-trained model. Second, our future work will extend to downstream tasks based on the restored SPET images, such as disease diagnosis and medical image segmentation. Finally, given the time and cost constraints associated with multi-modality acquisition, our forthcoming studies will also include research on missing modalities and unpaired data.

\vskip 2mm
\noindent
\textbf{Acknowledgment}
\vskip 2mm
\noindent
This work was supported by the National Natural Science Foundation of China (62331006, 62505006), and the Fundamental Research Funds for the Central Universities.


 \bibliographystyle{TSTbib}
 \bibliography{refs}
 
\begin{strip}
\end{strip}


\begin{biography}[figures/Authors/zhangyingkai]
\noindent
\textbf{Yingkai Zhang}\ \  received the B.S. degree in computer science in 2022 from the Beijing Institute of Technology, where he is currently working toward the Ph.D. degree in computer science and technology. His research interests include low-level vision, image processing, and computational photography.
\end{biography}
\vskip 5mm

\begin{biography}[figures/Authors/chenshuang]
\noindent
\textbf{Shuang Chen} received the B.S. degree in computer science in 2021 from Xidian University, and the M.S. degree in computer science in 2024 from the Beijing Institute of Technology. Her research interests include low-level vision, and image processing.
\end{biography}
\vskip 5mm

\begin{biography}[figures/Authors/tianye]
\noindent
\textbf{Ye Tian}  received the B.S. degree from School of Information Science and Engineering, Lanzhou University, Lanzhou, China, in 2017, the M.S. degree from Peking University, Beijing, China, in 2020, and the Ph.D. degree from the School of Information and Electronics, Beijing Institute of Technology, Beijing, China, in 2024. Her current research interests include deep learning, image processing, and computational imaging.
\end{biography}

\begin{biography}[figures/Authors/gaoyunyi]
\noindent
\textbf{Yunyi Gao}  received the B.S. degree from Beijing Institute of Technology, China in 2023, where he is currently working toward the M.S. degree in computer science and technology. His research interests focus on computer vision and image processing.
\end{biography}
\vskip 5mm

\begin{biography}[figures/Authors/jiangjianyong]
\noindent
\textbf{Jianyong Jiang} received the B.S. and M.S. degrees in Engineering Physics from Tsinghua University, Beijing, China, in 2010 and 2012, respectively. He obtained the Ph.D. degree in Nuclear Engineering and Management from the University of Tokyo, Tokyo, Japan, in 2015. He is currently an Associate Professor at Beijing Normal University. His research interests encompass nuclear medicine imaging, and artificial intelligence-based methods for image reconstruction.
\end{biography}

\begin{biography}[figures/Authors/fuying]
\noindent
\textbf{Ying Fu}  received the B.S. degree in electronic engineering from Xidian University, Xi'an, China, in 2009, the M.S. degree in automation from Tsinghua University, Beijing, China, in 2012, and the Ph.D. degree in information science and technology from the University of Tokyo, Tokyo, Japan, in 2015. She is currently a professor with the School of Computer Science and Technology, Beijing Institute of Technology. Her research interests include physics-based vision, image and video processing, and computational photography.
\end{biography}

%

\end{document}